# Context-Specific Independence in Bayesian Networks

**Craig Boutilier**
Dept. of Computer Science
University of British Columbia
Vancouver, BC V6T 1Z4
*cebly@cs.ubc.ca*

**Nir Friedman**
Dept. of Computer Science
Stanford University
Stanford, CA 94305-9010
*nir@cs.stanford.edu*

**Moises Goldszmidt**
SRI International
333 Ravenswood Way, EK329
Menlo Park, CA 94025
*moises@erg.sri.com*

**Daphne Koller**
Dept. of Computer Science
Stanford University
Stanford, CA 94305-9010
*koller@cs.stanford.edu*

## Abstract

Bayesian networks provide a language for qualitatively representing the conditional independence properties of a distribution. This allows a natural and compact representation of the distribution, eases knowledge acquisition, and supports effective inference algorithms. It is well-known, however, that there are certain independencies that we cannot capture qualitatively within the Bayesian network structure: independencies that hold only in certain *contexts*, i.e., given a specific assignment of values to certain variables. In this paper, we propose a formal notion of *context-specific independence (CSI)*, based on regularities in the conditional probability tables (CPTs) at a node. We present a technique, analogous to (and based on) d-separation, for determining when such independence holds in a given network. We then focus on a particular qualitative representation scheme—tree-structured CPTs—for capturing CSI. We suggest ways in which this representation can be used to support effective inference algorithms. In particular, we present a structural decomposition of the resulting network which can improve the performance of clustering algorithms, and an alternative algorithm based on cutset conditioning.

## 1 Introduction

The power of Bayesian Network (BN) representations of probability distributions lies in the efficient encoding of independence relations among random variables. These independencies are exploited to provide savings in the representation of a distribution, ease of knowledge acquisition and domain modeling, and computational savings in the inference process.[1] The objective of this paper is to increase this power by refining the BN representation to capture additional independence relations. In particular, we investigate how independence given certain variable *assignments* can be exploited in BNs in much the same way independence among variables is exploited in current BN representations and inference algorithms. We formally characterize this structured representation and catalog a number of the advantages it provides.

A BN is a directed acyclic graph where each node represents a random variable of interest and edges represent direct correlations between the variables. The absence of edges between variables denotes statements of independence. More precisely, we say that variables $Z$ and $Y$ are *independent* given a set of variables $X$ if $P(z \mid x, y) = P(z \mid x)$ for all values $x$, $y$ and $z$ of variables $X$, $Y$ and $Z$. A BN encodes the following statement of independence about each random variable: a variable is independent of its non-descendants in the network given the state of its parents [14]. For example, in the network shown in Figure 1, $Z$ is independent of $U$, $V$ and $Y$ given $X$ and $W$. Further independence statements that follow from these local statements can be read from the network structure, in polynomial time, using a graph-theoretic criterion called *d-separation* [14].

In addition to representing statements of independence, a BN also represents a particular distribution (that satisfies all the independencies). This distribution is specified by a set of *conditional probability tables* (CPTs). Each node $X$ has an associated CPT that describes the conditional distribution of $X$ given different assignments of values for its parents. Using the independencies encoded in the structure of the network, the joint distribution can be computed by simply multiplying the CPTs.

In its most naive form, a CPT is encoded using a tabular representation in which each assignment of values to the parents of $X$ requires the specification of a conditional distribution over $X$. Thus, for example, assuming that all of $U$, $V$, $W$ and $X$ in Figure 1 are binary, we need to specify eight such distributions (or eight parameters). The size of this representation is exponential in the number of parents. Furthermore, this representation fails to capture certain regularities in the node distribution. In the CPT of Figure 1, for example, $P(x \mid u, V, W)$ is equal to some constant $p_1$ regardless of the values taken by $V$ and $W$: when $u$ holds (i.e., when $U = t$) we need not consider the values of $V$ and $W$. Clearly, we need to specify at most five distributions over $X$ instead of eight. Such regularities occur often enough that at least two well known BN products—Microsoft's Bayesian Networks Modeling Tool and Knowledge Industries' DXpress—have incorporated special mechanisms in their knowledge acquisition interface that allow the user to more easily specify the corresponding CPTs.

---

[1] Inference refers to the computation of a posterior distribution, conditioned on evidence.

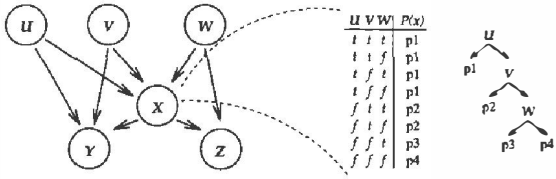

Figure 1: Context-Specific Independence

In this paper, we provide a formal foundation for such regularities by using the notion of *context-specific independence*. Intuitively, in our example, the regularities in the CPT of $X$ ensure that $X$ is independent of $W$ and $V$ given the *context u* ($U = t$), but is dependent on $W, V$ in the context $\overline{u}$ ($U = f$). This is an assertion of context-specific independence (CSI), which is more restricted than the statements of *variable independence* that are encoded by the BN structure. Nevertheless, as we show in this paper, such statements can be used to extend the advantages of variable independence for probabilistic inference, namely, ease of knowledge elicitation, compact representation and computational benefits in inference.

We are certainly not the first to suggest extensions to the BN representation in order to capture additional independencies and (potentially) enhance inference. Well-known examples include Heckerman's [9] *similarity networks* (and the related *multinets* [7]), the use of asymmetric representations for decision making [18, 6] and Poole's [16] use of probabilistic Horn rules to encode dependencies between variables. Even the representation we emphasize (decision trees) have been used to encode CPTs [2, 8]. The intent of this work is to formalize the notion of CSI, to study its representation as part of a more general framework, and to propose methods for utilizing these representations to enhance probabilistic inference algorithms.

We begin in Section 2 by defining context-specific independence formally, and introducing a simple, local transformation for a BN based on arc deletion so that CSI statements can be readily determined using d-separation. Section 3 discusses in detail how trees can be used to represent CPTs compactly, and how this representation can be exploited by the algorithms for determining CSI. Section 4 offers suggestions for speeding up probabilistic inference by taking advantage of CSI. We present network transformations that may reduce clique size for clustering algorithms, as well as techniques that use CSI—and the associated arc-deletion strategy—in cutset conditioning. We conclude with a discussion of related notions and future research directions.

## 2 Context-Specific Independence and Arc Deletion

Consider a finite set $U = \{X_1, \ldots, X_n\}$ of discrete random variables where each variable $X_i \in U$ may take on values from a finite domain. We use capital letters, such as $X, Y, Z$, for variable names and lowercase letters $x, y, z$ to denote specific values taken by those variables. The set of all values of $X$ is denoted $val(X)$. Sets of variables are denoted by boldface capital letters $\mathbf{X}, \mathbf{Y}, \mathbf{Z}$, and assignments of values to the variables in these sets will be denoted by boldface lowercase letters $\mathbf{x}, \mathbf{y}, \mathbf{z}$ (we use $val(\mathbf{X})$ in the obvious way).

**Definition 2.1:** Let $P$ be a joint probability distribution over the variables in $U$, and let $\mathbf{X}, \mathbf{Y}, \mathbf{Z}$ be subsets of $U$. $\mathbf{X}$ and $\mathbf{Y}$ are *conditionally independent* given $\mathbf{Z}$, denoted $I(\mathbf{X}; \mathbf{Y} \mid \mathbf{Z})$, if for all $\mathbf{x} \in val(\mathbf{X}), \mathbf{y} \in val(\mathbf{Y}), \mathbf{z} \in val(\mathbf{Z})$, the following relationship holds:

$$P(\mathbf{x} \mid \mathbf{z}, \mathbf{y}) = P(\mathbf{x} \mid \mathbf{z}) \text{ whenever } P(\mathbf{y}, \mathbf{z}) > 0. \quad (1)$$

We summarize this last statement (for all values of $\mathbf{x}, \mathbf{y}, \mathbf{z}$) by $P(\mathbf{X} \mid \mathbf{Z}, \mathbf{Y}) = P(\mathbf{X} \mid \mathbf{Z})$.

A *Bayesian network* is a directed acyclic graph $B$ whose nodes correspond to the random variables $X_1, \ldots, X_n$, and whose edges represent direct dependencies between the variables. The graph structure of $B$ encodes the set of independence assumptions representing the assertion that each node $X_i$ is independent of its non-descendants given its parents $\Pi_{X_i}$. These statements are *local*, in that they involve only a node and its parents in $B$. Other $I()$ statements, involving arbitrary sets of variables, follow from these local assertions. These can be read from the structure of $B$ using a graph-theoretic path criterion called *d-separation* [14] that can be tested in polynomial time.

A BN $B$ represents independence information about a particular distribution $P$. Thus, we require that the independencies encoded in $B$ hold for $P$. More precisely, $B$ is said to be an *I-map* for the distribution $P$ if every independence sanctioned by d-separation in $B$ holds in $P$. A BN is required to be a *minimal* I-map, in the sense that the deletion of any edge in the network destroys the I-mapness of the network with respect to the distribution it describes. A BN $B$ for $P$ permits a compact representation of the distribution: we need only specify, for each variable $X_i$, a *conditional probability table (CPT)* encoding a parameter $P(x_i \mid \Pi_{x_i})$ for each possible value of the variables in $\{X_i, \Pi_{X_i}\}$. (See [14] for details.)

The graphical structure of the BN can only capture independence relations of the form $I(\mathbf{X}; \mathbf{Y} \mid \mathbf{Z})$, that is, independencies that hold for any assignment of values to the variables in $\mathbf{Z}$. However, we are often interested in independencies that hold only in certain contexts.

**Definition 2.2:** Let $\mathbf{X}, \mathbf{Y}, \mathbf{Z}, \mathbf{C}$ be pairwise disjoint sets of variables. $\mathbf{X}$ and $\mathbf{Y}$ are *contextually independent* given

$Z$ and the *context* $c \in val(C)$, denoted $I_c(X;Y \mid Z,c)$, if

$$P(X \mid Z,c,Y) = P(X \mid Z,c) \text{ whenever } P(Y,Z,c) > 0.$$

This assertion is similar to that in Equation (1), taking $C \cup Z$ as evidence, but requires that the independence of $X$ and $Y$ hold only for the particular assignment $c$ to $C$.

It is easy to see that certain *local* $I_c$ statements — those of the form $I_c(X;Y \mid c)$ for $Y, C \subseteq \Pi_X$ — can be verified by direct examination of the CPT for $X$. In Figure 1, for example, we can verify $I_c(X;V \mid u)$ by checking in the CPT for $X$ whether, for each value $w$ of $W$, $P(X \mid v,w,u)$ does not depend on $v$ (i.e., it is the same for all values $v$ of $V$). The next section explores different representations of the CPTs that will allow us to check these local statements efficiently. Our objective now is to establish an analogue to the principle of d-separation: a computationally tractable method for deciding the validity of non-local $I_c$ statements. It turns out that this problem can be solved by a simple reduction to a problem of validating variable independence statements in a simpler network. The latter problem can be efficiently solved using $d$-separation.

**Definition 2.3:** An edge from $Y$ into $X$ will be called *vacuous* in $B$, given a context $c$, if $I_c(X;Y \mid c \cap \Pi_X)$. Given BN $B$ and a context $c$, we define $B(c)$ as the BN that results from deleting vacuous edges in $B$ given $c$. We say that $X$ is *CSI-separated* from $Y$ given $Z$ in context $c$ in $B$ if $X$ is d-separated from $Y$ given $Z \cup C$ in $B(c)$.

Note that the statement $I_c(X;Y \mid c \cap \Pi_X)$ is a local $I_c$ statement and can be determined by inspecting the CPT for $X$. Thus, we can decide CSI-separation by transforming $B$ into $B(c)$ using these local $I_c$ statements to delete vacuous edges, and then using d-separation on the resulting network.

We now show that this notion of CSI-separation is sound and (in a strong sense) complete given these local independence statements. Let $B$ be a network structure and $\mathcal{I}_c^\ell$ be a set of local $I_c$ statements over $B$. We say that $(B, \mathcal{I}_c^\ell)$ is a *CSI-map* of a distribution $P$ if the independencies implied by $(B, \mathcal{I}_c^\ell)$ hold in $P$, i.e., $I_c(X;Y \mid Z,c)$ holds in $P$ whenever $X$ is *CSI-separated* from $Y$ given $Z$ in context $c$ in $(B, \mathcal{I}_c^\ell)$. We say that $(B, \mathcal{I}_c^\ell)$ is a *perfect CSI-map* if the implied independencies are the only ones that hold in $P$, i.e., if $I_c(X;Y \mid Z,c)$ if and only if $X$ is *CSI-separated* from $Y$ given $Z$ in context $c$ in $(B, \mathcal{I}_c^\ell)$

**Theorem 2.4:** *Let $B$ be a network structure, $\mathcal{I}_c^\ell$ be a set of local independencies, and $P$ a distribution consistent with $B$ and $\mathcal{I}_c^\ell$. Then $(B, \mathcal{I}_c^\ell)$ is a CSI-map of $P$.*

The theorem establishes the soundness of this procedure. Is the procedure also complete? As for any such procedure, there may be independencies that we cannot detect using only local independencies and network structure. However, the following theorem shows that, in a sense, this procedure provides the best results that we can hope to derive based solely on the structural properties of the distribution.

**Theorem 2.5:** *Let $B$ be a network structure, $\mathcal{I}_c^\ell$ be a set of local independencies. Then there exists a distribution $P$, consistent with $B$ and $\mathcal{I}_c^\ell$, such that $(B, \mathcal{I}_c^\ell)$ is a perfect CSI-map of $P$.*

## 3  Structured Representations of CPTs

Context-specific independence corresponds to regularities within CPTs. In this section, we discuss possible representations that capture this regularity qualitatively, in much the same way that a BN structure qualitatively captures conditional independence. Such representations admit effective algorithms for determining local CSI statements and can be exploited in probabilistic inference. For reasons of space, we focus primarily on tree-structured representations.

In general, we can view a CPT as a function that maps $val(\Pi_X)$ into distributions over $X$. A compact representation of CPTs is simply a representation of this function that exploits the fact that distinct elements of $val(\Pi_X)$ are associated with the same distribution. Therefore, one can compactly represent CPTs by simply partitioning the space $val(\Pi_X)$ into regions mapping to the same distribution.

Most generally, we can represent the partitions using a set of mutually exclusive and exhaustive generalized propositions over the variable set $\Pi_X$. A generalized proposition is simply a truth functional combination of specific variable assignments, so that if $Y, Z \in \Pi_X$, we may have a partition characterized by the generalized proposition $(Y = y) \vee \neg(Z = z)$. Each such proposition is associated with a distribution over $X$. While this representation is fully general, it does not easily support either probabilistic inference or inference about CSI. Fortunately, we can often use other, more convenient, representations for this type of partitioning. For example, one could use a canonical logical form such as minimal CNF. Classification trees (also known in the machine learning community as decision trees) are another popular function representation, with partitions of the state space induced by the labeling of branches in the tree. These representations have a number of advantages, including the fact that vacuous edges can be detected, and reduced CPTs produced in linear time (in the size of the CPT representation). As expected, there is a tradeoff: the most compact CNF or tree representation of a CPT might be much larger (exponentially larger in the worst case) than the minimal representation in terms of generalized propositions.

For the purposes of this paper, we focus on *CPT-trees*—tree-structured CPTs, deferring discussion of analogous results for CNF representations and graph-structured CPTs (of the form discussed by [3]) to a longer version of this paper. A major advantage of tree structures is their naturalness, with branch labels corresponding in some sense to "rule" structure (see Figure 1). This intuition makes it particularly easy to elicit probabilities directly from a human expert. As we show in subsequent sections, the tree structure can also be utilized to speed up BN inference algorithms. Finally, as we discuss in the conclusion, trees are also amenable to well-studied approximation and learning

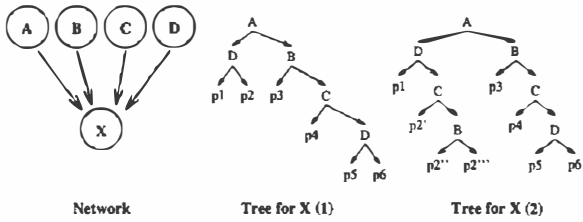

Figure 2: CPT-tree Representation

methods [17]. In this section, we show that they admit fast algorithms for detecting CSI.

In general, there are two operations we wish to perform given a context $c$: the first is to determine whether a given arc into a variable $X$ is vacuous; the second is to determine a *reduced CPT* when we condition on $c$. This operation is carried out whenever we set evidence and should reflect the changes to $X$'s parents that are implied by context-specific independencies given $c$. We examine how to perform both types of operations on CPT-trees. To avoid confusion, we use t-node and t-arc to denote nodes and arcs in the tree (as opposed to nodes and arcs in the BN). To illustrate these ideas, consider the CPT-tree for the variable $X$ in Figure 2. (Left t-arcs are labeled true and right t-arcs false).

Given this representation, it is relatively easy to tell which parents are rendered independent of $X$ given context $c$. Assume that Tree 1 represents the CPT for $X$. In context $a$, clearly $D$ remains relevant while $C$ and $B$ are rendered independent of $X$. Given $\bar{a} \wedge b$, both $C$ and $D$ are rendered independent of $X$. Intuitively, this is so because the distribution on $X$ does not depend on $C$ and $D$ once we know $c = \bar{a} \wedge b$: every path from the root to leaf which is consistent with $c$ fails to mention $C$ or $D$.

**Definition 3.1:** A *path* in the CPT-tree is the set of t-arcs lying between the root and a leaf. The *labeling* of a path is the assignment to variables induced by the labels on the t-arcs of the path. A variable $Y$ *occurs* on a path if one of the t-nodes along the path tests the value of $Y$. A path is *consistent* with a context $c$ iff the labeling of the path is consistent with the assignment of values in $c$.

**Theorem 3.2:** *Let $T$ be a CPT-tree for $X$ and let $Y$ be one of its parents. Let $c \in C$ be some context ($Y \notin C$). If $Y$ does not lie on any path consistent with $c$, then the edge $Y \to X$ is vacuous given $c$.*

This provides us with a *sound* test for context-specific independence (only valid independencies are discovered). However, the test is not *complete*, since there are CPT structures that cannot be represented minimally by a tree. For instance, suppose that $p1 = p5$ and $p2 = p6$ in the example above. Given context $\bar{b} \wedge \bar{c}$, we can tell that $A$ is irrelevant by inspection; but, the choice of variable ordering prevents us from detecting this using the criterion in the theorem. However, the test above is complete in the sense that no other edge is vacuous given the tree structure.

**Theorem 3.3:** *Let $T$ be a CPT-tree for $X$, let $Y \in \Pi_X$ and let $c \in C$ be some context ($Y \notin C$). If $Y$ occurs on a path that is consistent with $c$, then there exists an assignment of parameters to the leaves of $T$ such that $Y \to X$ is not vacuous given $c$.*

This shows that the test described above is, in fact, the best test that uses only the structure of the tree and not the actual probabilities. This is similar in spirit to d-separation: it detects all conditional independencies possible from the structure of the network, but it cannot detect independencies that are hidden in the quantification of the links. As for conditional independence in belief networks, we need only soundness in order to exploit CSI in inference.

It is also straightforward to produce a reduced CPT-tree representing the CPT conditioned on context $c$. Assume $c$ an assignment to variables containing certain parents of $X$ and $T$ is the CPT-tree of $X$, with root $R$ and immediate subtrees $T_1, \cdots T_k$. The *reduced CPT-tree* $T(c)$ is defined recursively as follows: if the label of $R$ is not among the variables $C$, then $T(c)$ consists of $R$ with subtrees $T_j(c)$; if the label of $R$ is some $Y \in C$, then $T(c) = T_j(c)$, where $T_j$ is the subtree pointed to by the t-arc labeled with value $y \in c$. Thus, the reduced tree $T(c)$ can be produced with one tree traversal in $O(|T|)$ time.

**Proposition 3.4:** *Variable $Y$ labels some t-node in $T(c)$ if and only if $Y \notin C$ and $Y$ occurs on a path in $T$ that is consistent with $c$.*

This implies that $Y$ appears in $T(c)$ if and only if $Y \to X$ is not deemed vacuous by the test described above. Given the reduced tree, determining the list of arcs pointing into $X$ that can be deleted requires a simple tree traversal of $T(c)$. Thus, reducing the tree gives us an efficient and sound test for determining the context-specific independence of all parents of $X$.

## 4   Exploiting CSI in Probabilistic Inference

Network representations of distributions offer considerable computational advantages in probabilistic inference. The graphical structure of a BN lays bare variable independence relationships that are exploited by well-known algorithms when deciding what information is relevant to (say) a given query, and how best that information can be summarized. In a similar fashion, compact representations of CPTs such as trees make CSI relationships explicit. In this section, we describe how CSI might be exploited in various BN inference algorithms, specifically stressing particular uses in clustering and cutset conditioning. Space precludes a detailed presentation; we provide only the basic intuitions here. We also emphasize that these are by no means the only ways in which BN inference can employ CSI.

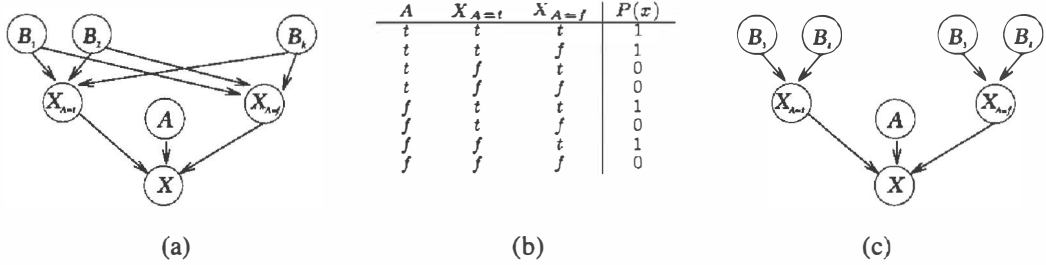

Figure 3: (a) A simple decomposition of the node $X$; (b) The CPT for the new node $X$; (c) A more effective decomposition of $X$, utilizing CSI.

### 4.1 Network Transformations and Clustering

The use of compact representations for CPTs is not a novel idea. For instance, noisy-or distributions (or generalizations [19]) allow compact representation by assuming that the parents of $X$ make independent "casual contributions" to the value of $X$. These distributions fall into the general category of distributions satisfying causal independence [10, 11]. For such distributions, we can perform a structural transformation on our original network, resulting in a new network where many of these independencies are encoded qualitatively *within the network structure*. Essentially, the transformation introduces auxiliary variables into the network, then connects them via a cascading sequence of deterministic or-nodes [11]. While CSI is quite distinct from causal independence, similar ideas can be applied: a structural network transformation can be used to capture certain aspects of CSI directly within the BN-structure.

Such transformations can be very useful when one uses an inference algorithm based on clustering [13]. Roughly speaking, clustering algorithms construct a *join tree*, whose nodes denote (overlapping) clusters of variables in the original BN. Each cluster, or *clique*, encodes the marginal distribution over the set $val(\boldsymbol{X})$ of the nodes $\boldsymbol{X}$ in the cluster. The inference process is carried out on the join tree, and its complexity is determined largely by the size of the largest clique. This is where the structural transformations prove worthwhile. The clustering process requires that each family in the BN — a node and its parents — be a subset of at least one clique in the join tree. Therefore, a family with a large set of values $val(\{X_i\} \cup \Pi_{X_i})$ will lead to a large clique and thereby to poor performance of clustering algorithms. A transformation that reduces the overall number of values present in a family can offer considerable computational savings in clustering algorithms.

In order to understand our transformation, we first consider a generic node $X$ in a Bayesian network. Let $A$ be one of $X$'s parents, and let $B_1, \ldots, B_k$ be the remaining parents. Assume, for simplicity, that $X$ and $A$ are both binary-valued. Intuitively, we can view the value of the random variable $X$ as the outcome of two conditional variables: the value that $X$ would take if $A$ were true, and the value that $X$ would take if $A$ were false. We can conduct a thought experiment where these two variables are decided separately, and then, when the value of $A$ is revealed, the appropriate value for $X$ is chosen.

Formally, we define a random variable $X_{A=t}$, with a conditional distribution that depends only on $B_1, \ldots, B_k$:

$$P(X_{A=t} \mid B_1, \ldots, B_k) = P(X \mid A = t, B_1, \ldots, B_k)$$

We can similarly define a variable $X_{A=f}$. The variable $X$ is equal to $X_{A=t}$ if $A = t$ and is equal to $X_{A=f}$ if $A = f$. Note that the variables $X_{A=t}$ and $X_{A=f}$ both have the same set of values as $X$. This perspective allows us to replace the node $X$ in any network with the subnetwork illustrated in Figure 3(a). The node $X$ is a deterministic node, which we call a *multiplexer node* (since $X$ takes either the value of $X_{A=t}$ or of $X_{A=f}$, depending on the value of $A$). Its CPT is presented in Figure 3(b).

For a generic node $X$, this decomposition is not particularly useful. For one thing, the total size of the two new CPTs is exactly the same as the size of the original CPT for $X$; for another, the resulting structure (with its many tightly-coupled cycles) does not admit a more effective decompositions into cliques. However, if $X$ exhibits a significant amount of CSI, this type of transformation can result in a far more compact representation. For example, let $k = 4$, and assume that $X$ depends only on $B_1$ and $B_2$ when $A$ is true, and only on $B_3$ and $B_4$ when $A$ is false. Then each of $X_{A=t}$ and $X_{A=f}$ will have only two parents, as in Figure 3(c). If these variables are binary, the new representation requires two CPTs with four entries each, plus a single deterministic multiplexer node with 8 (predetermined) 'distributions'. By contrast, the original representation of $X$ had a single CPT with 32 entries. Furthermore, the structure of the resulting network may well allow the construction of a join tree with much smaller cliques.

Our transformation uses the structure of a CPT-tree to apply this decomposition recursively. Essentially, each node $X$ is first decomposed according to the parent $A$ which is at the root of its CPT tree. Each of the conditional nodes ($X_{A=t}$ and $X_{A=f}$ in the binary case) has, as its CPT, one of the subtrees of the t-node $A$ in the CPT for $X$. The resulting conditional nodes can be decomposed recursively, in a similar fashion. In Figure 4, for example, the node corresponding to $X_{A=f}$ can be decomposed into $X_{A=f,B=t}$ and $X_{A=f,B=f}$. The node $X_{A=f,B=f}$ can then be decomposed

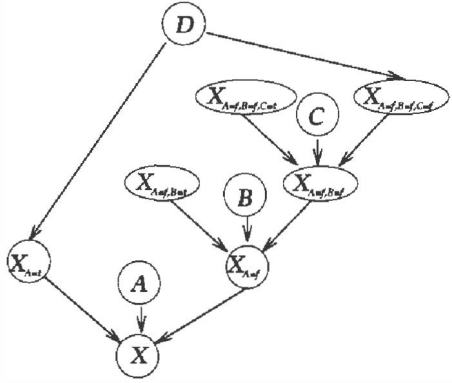

Figure 4: A decomposition of the network in Figure 2, according to Tree (1).

into $X_{A=f,B=f,C=t}$ and $X_{A=f,B=f,C=f}$.

The nodes $X_{A=f,B=t}$ and $X_{A=f,B=f,C=t}$ cannot be decomposed further, since they have no parents. While further decomposition of nodes $X_{A=t}$ and $X_{A=f,B=f,C=f}$ is possible, this is not beneficial, since the CPTs for these nodes are unstructured (a complete tree of depth 1). It is clear that this procedure is beneficial only if there is a structure in the CPT of a node. Thus, in general, we want to stop the decomposition when the CPT of a node is a full tree. (Note that this includes leaves a special case.)

As in the structural transformation for noisy-or nodes of [11], our decomposition can allow clustering algorithms to form smaller cliques. After the transformation, we have many more nodes in the network (on the order of the size of all CPT tree representations), but each generally has far fewer parents. For example, Figure 4 describes the transformation of the CPT of Tree (1) of Figure 2. In this transformation we have eliminated a family with four parents and introduced several smaller families. We are currently working on implementing these ideas, and testing their effectiveness in practice. We also note that a large fraction of the auxiliary nodes we introduce are multiplexer nodes, which are deterministic function of their parents. Such nodes can be further exploited in the clustering algorithm [12].

We note that the reduction in clique size (and the resulting computational savings) depend heavily on the structure of the decision trees. A similar phenomenon occurs in the transformation of [11], where the effectiveness depends on the order in which we choose to cascade the different parents of the node.

As in the case of noisy-or, the graphical structure of our (transformed) BN cannot capture all independencies implicit in the CPTs. In particular, none of the CSI relations—induced by particular value assignments—can be read from the transformed structure. In the noisy-or case, the analogue is our inability to structurally represent that a node's parents are independent if the node is observed to be false, but not if it is observed to be true.[2] In both cases, these CSI relations are captured by the deterministic relationships used in the transformation: in an "or" node, the parents are independent if the node is set to false. In a multiplexer node, the value depends only on one parent once the value of the "selecting" parent (the original variable) is known.

### 4.2 Cutset Conditioning

Even using noisy-or or tree representations, the join-tree algorithm can only take advantage of fixed structural independencies. The use of *static precompilation* makes it difficult for the algorithm to take advantage of independencies that only occur in certain circumstances, e.g., as new evidence arrives. More dynamic algorithms, such as cutset conditioning [14], can exploit context-specific independencies more effectively. We investigate below how cutset algorithms can be modified to exploit CSI using our decision-tree representation.[3]

The cutset conditioning algorithm works roughly as follows. We select a *cutset*, i.e., a set of variables that, once instantiated, render the network singly connected. Inference is then carried out using reasoning by cases, where each case is a possible assignment to the variables in the cutset $C$. Each such assignment is instantiated as evidence in a call to the *polytree* algorithm [14], which performs inference on the resulting network. The results of these calls are combined to give the final answer. The running time is largely determined by the number of calls to the polytree algorithm (i.e., $|val(C)|$).

CSI offers a rather obvious advantage to inference algorithms based on the conditioning of loop cutsets. By instantiating a particular variable to a certain value in order to cut a loop, CSI may render other arcs vacuous, perhaps cutting additional loops without the need for instantiating additional variables. For instance, suppose the network in Figure 1 is to be solved using the cutset $\{U, V, W\}$ (this might be the optimal strategy if $|val(X)|$ is very large). Typically, we solve the reduced singly-connected network $|val(U)| \cdot |val(V)| \cdot |val(W)|$ times, once for each assignment of values to $U, V, W$. However, by recognizing the fact that the connections between $X$ and $\{V, W\}$ are vacuous in context $u$, we need not instantiate $V$ and $W$ when we assign $U = t$. This replaces $|val(V)| \cdot |val(W)|$ network evaluations with a single evaluation. However, when $U = f$, the instantiation of $V, W$ can no longer be ignored (the edges are not vacuous in context $\overline{u}$).

To capture this phenomenon, we generalize the standard notion of a cutset by considering tree representations of cutsets. These reflect the need to instantiate certain variables in some contexts, but not in others, in order to render the network singly-connected. Intuitively, a *conditional cutset* is a tree with interior nodes labeled by variables and edges la-beled by (sets of) variable values.[4] Each branch through the tree corresponds to the set of assignments induced by fixing one variable value on each edge. The tree is a conditional cutset if: (a) each branch through the tree represents a context that renders the network singly-connected; and (b) the set of such assignments is mutually exclusive and exhaustive. Examples of conditional cutsets for the BN in Figure 1 are illustrated in Figure 5: (a) is the obvious compact cutset; (b) is the tree representation of the "standard" cutset, which fails to exploit the structure of the CPT, requiring one evaluation for each instantiation of $U, V, W$.

---

[2] This last fact is heavily utilized by algorithms targeted specifically at noisy-or networks (mostly BN2O networks).

[3] We believe similar ideas can be applied to other compact CPT representations such as noisy-or.

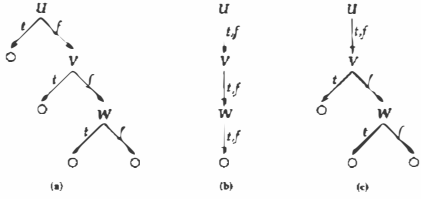

Figure 5: Valid Conditional Cutsets

Once we have a conditional cutset in hand, the extension of classical cutset inference is fairly obvious. We consider each assignment of values to variables determined by branches through the tree, instantiate the network with this assignment, run the polytree algorithm on the resulting network, and combine the results as usual.[5] Clearly, the complexity of this algorithm is a function of the number of distinct paths through the conditional cutset. It is therefore crucial to find good heuristic algorithms for constructing small conditional cutsets. We focus on a "computationally intensive" heuristic approach that exploits CSI and the existence of vacuous arcs maximally. This algorithm constructs conditional cutsets incrementally, in a fashion similar to standard heuristic approaches to the problem [20, 1]. We discuss computationally-motivated shortcuts near the end of this section.

The standard "greedy" approach to cutset construction selects nodes for the cutset according to the heuristic value $\frac{w(X)}{d(X)}$, where the *weight* $w(X)$ of variable $X$ is $\log(|val(X)|)$ and $d(X)$ is the out-degree of $X$ in the network graph [20, 1].[6] The weight measures the work needed to instantiate $X$ in a cutset, while the degree of a vertex gives an idea of its arc-cutting potential—more incident outgoing edges mean a larger chance to cut loops. In order to extend this heuristic to deal with CSI, we must estimate the extent to which arcs are cut due to CSI. The obvious approach, namely adding to $d(X)$ the number of arcs *actually* rendered vacuous by $X$ (averaging over values of $X$), is reasonably straightforward, but unfortunately is somewhat myopic. In particular, it ignores the *potential* for arcs to be cut subsequently. For example, consider the family in Figure 2, with Tree 2 reflecting the CPT for $X$. Adding $A$ or $B$ to a cutset causes no additional arcs into $X$ to be cut, so they will have the same heuristic value (other things being equal). However, clearly $A$ is the more desirable choice because, given either value of $A$, the conditional cutsets produced *subsequently* using $B$, $C$ and $D$ will be very small.

Rather than using the *actual* number of arcs cut by selecting a node for the cutset, we should consider the *expected* number of arcs that will be cut. We do this by considering, for each of the children $V$ of $X$, how many distinct probability entries (distributions) are found in the structured representation of the CPT for that child for each instantiation $X = x_i$ (i.e., the size of the *reduced* CPT). The log of this value is the expected number of parents required for the child $V$ after $X = x_i$ is known, with fewer parents indicating more potential for arc-cutting. We can then average this number for each of the values $X$ may take, and sum the expected number of cut arcs for each of $X$'s children. This measure then plays the role of $d(X)$ in the cutset heuristic. More precisely, let $t(V)$ be the size of the CPT-structure (i.e., number of entries) for $V$ in a fixed network; and let $t(V, x_i)$ be the size of the reduced CPT given context $X = x_i$ (we assume $X$ is a parent of $V$). We define the *expected number of parents* of $V$ given $x_i$ to be

$$EP(V, x_i) = \frac{\sum_{A \in Parents(V) - X} \log_{|val(A)|} t(V, X = x_i)}{|Parents(V)| - 1}$$

The *expected number of arc deletions* from $B$ if $X$ is instantiated is given by

$$d'(X) = \frac{\sum_{V \in Children(X)} \sum_{x_i \in val(X)} |Parents(V)| - EP(V, x_i)}{|val(X)|}$$

Thus, $\frac{w(X)}{d'(X)}$ gives an reasonably accurate picture of the value of adding $X$ to a conditional cutset in a network $B$.

Our cutset construction algorithm proceeds recursively by: 1) adding a heuristically selected node $X$ to a branch of the tree-structured cutset; 2) adding t-arcs to the cutset-tree for each value $x_i \in val(X)$; 3) constructing a new network for each of these instantiations of $X$ that reflects CSI; and 4) extending each of these new arcs recursively by selecting the node that looks best in the new network corresponding to that branch. We can very roughly sketch it as follows. The algorithm begins with the original network $B$.

1. Remove singly-connected nodes from $B$, leaving $B_r$. If no nodes remain, return the empty cutset-tree.

2. Choose node $X$ in $B_r$ s.t. $w(X)/d'(X)$ is minimal.

3. For each $x_i \in val(X)$, construct $B_{x_i}$ by removing vacuous arcs from $B_r$ and replacing all CPTs by the reduced CPTs using $X = x_i$.

4. Return the tree $T'$ where: a) $X$ labels the root of $T'$; b) one t-arc for each $x_i$ emanates from the root; and c)

---

[4] We explain the need for set-valued arc labels below.

[5] As in the standard cutset algorithm, the weights required to combine the answers from the different cases can be obtained from the polytree computations [21].

[6] We assume that the network has been preprocessed by node-splitting so that legitimate cutsets can be selected easily. See [1] for details.

the t-node attached to the end of the $x_i$ t-arc is the tree produced by recursively calling the algorithm with the network $B_{x_i}$.

Step 1 of the algorithm is standard [20, 1]. In Step 2, it is important to realize that the heuristic value of $X$ is determined with respect to the current network and the context already established in the existing branch of the cutset. Step 3 is required to ensure that the selection of the next variable reflects the fact that $X = x_i$ is part of the current context. Finally, Step 4 emphasizes the conditional nature of variable selection: different variables may be selected next given different values of $X$. Steps 2–4 capture the key features of our approach and have certain computational implications, to which we now turn our attention.

Our algorithm exploits CSI to a great degree, but requires computational effort greater than that for normal cutset construction. First, the cutset itself is structured: a tree representation of a standard cutset is potentially exponentially larger (a full tree). However, the algorithm can be run on-line, and the tree never completely stored: as variables are instantiated to particular values for conditioning, the selection of the next variable can be made. Conceptually, this amounts to a depth-first construction of the tree, with only one (partial or complete) branch ever being stored. In addition, we can add an optional step before Step 4 that detects structural equivalence in the networks $B_{x_i}$. If, say, the instantiations of $X$ to $x_i$ and $x_j$ have the same structural effect on the arcs in $B$ and the representation of reduced CPTs, then we need not distinguish these instantiations subsequently (in cutset construction). Rather, in Step 4, we would create one new t-arc in the cutset-tree labeled with the set $\{x_i, x_j\}$ (as in Figure 5). This reduces the number of graphs that need to be constructed (and concomitant computations discussed below). In completely unstructured settings, the representation of a conditional cutset would be of size similar to a normal cutset, as in Figure 5(b).

Apart from the amount of information in a conditional cutset, more effort is needed to decide which variables to add to a branch, since the heuristic component $d'(X)$ is more involved than vertex degree. Unfortunately, the value $d'(X)$ is not fixed (in which case it would involve a single set of prior computations); it must be recomputed in Step 2 to reflect the variable instantiations that gave rise to the current network. Part of the re-evaluation of $d'(X)$ requires that CPTs also be updated (Step 3). Fortunately, the number of CPTs that have to be updated for assignment $X = x_i$ is small: only the children of $X$ (in the current graph) need to have CPTs updated. This can be done using the CPT reduction algorithms described above, which are very efficient. These updates then affect the heuristic estimates of only their parents; i.e., only the "spouses" $V$ of $X$ need to have their value $d'(V)$ recomputed. Thus, the amount of work required is not too large, so that the reduction in the number of network evaluations will usually compensate for the extra work. We are currently in the process of implementing this algorithm to test its performance in practice.

There are several other directions that we are currently investigating in order to enhance this algorithm. One involves developing less ideal but more tractable methods of conditional cutset construction. For example, we might select a cutset by standard means, and use the considerations described above to order (on-line) the variable instantiations within this cutset. Another direction involves integrating these ideas with the computation-saving ideas of [4] for standard cutset algorithms.

## 5 Concluding Remarks

We have defined the notion of context-specific independence as a way of capturing the independencies induced by specific variable assignments, adding to the regularities in distributions representable in BNs. Our results provide foundations for CSI, its representation and its role in inference. In particular, we have shown how CSI can be determined using local computation in a BN and how specific mechanisms (in particular, trees) allow compact representation of CPTs and enable efficient detection of CSI. Furthermore, CSI and tree-structured CPTs can be used to speed up probabilistic inference in both clustering and cutset-style algorithms.

As we mentioned in the introduction, there has been considerable work on extending the BN representation to capture additional independencies. Our notion of CSI is related to what Heckerman calls *subset independence* in his work on similarity networks [9]. Yet, our approach is significantly different in that we try to capture the additional independencies by providing a structured representation of the CPTs within a single network, while similarity networks and multinets [9, 7] rely on a family of networks. In fact the approach we described based on decision trees is closer in spirit to that of Poole's rule-based representations of networks [16].

The arc-cutting technique and network transformation introduced in Section 2 is reminiscent of the network transformations introduced by Pearl in his probabilistic calculus of action [15]. Indeed the semantics of actions proposed in that paper can be viewed as an instance of CSI. This is not a mere coincidence, as it is easy to see that networks representing plans and influence diagrams usually contain a significant amount of CSI. The effects of actions (or decisions) usually only take place for specific instantiation of some variables, and are vacuous or trivial when these instantiations are not realized. Testimony to this fact is the work on adding additional structure to influence diagrams by Smith et al. [18], Fung and Shachter [6], and the work by Boutilier et al [2] on using decision trees to represent CPTs in the context of Markov Decision Processes.

There are a number of future research directions that are needed to elaborate the ideas presented here, and to expand the role that CSI and compact CPT representations play in probabilistic reasoning. We are currently exploring the use of different CPT representations, such as decision graphs, and the potential interaction between CSI and causal inde-pendence (as in the noisy-or model). A deeper examination of the network transformation algorithm of Section 4.1 and empirical experiments are necessary to determine the circumstances under which the reductions in clique size are significant. Similar studies are being conducted for the conditional cutset algorithm of Section 4.2 (and its variants). In particular, to determine the extent of the overhead involved in conditional cutset construction. We are currently pursuing both of these directions.

CSI can also play a significant role in approximate probabilistic inference. In many cases, we may wish to trade a certain amount of accuracy to speed up inference, allow more compact representation or ease knowledge acquisition. For instance, a CPT exhibiting little structure (i.e., little or no CSI) cannot be compactly represented; e.g., it may require a full tree. However, in many cases, the local dependence is weaker in some circumstances than in others. Consider Tree 2 in Figure 2 and suppose that none of $p2', p2'', p2'''$ are very different, reflecting the fact the influence of $B$ and $C$'s on $X$ is relatively weak in the case where $A$ is true and $D$ is false. In this case, we may assume that these three entries are actually the same, thus approximating the true CPT using a decision tree with the structure of Tree 1.

This saving (both in representation and inference, using the techniques of this paper) comes at the expense of accuracy. In ongoing work, we show how to estimate the (cross-entropy) error of a local approximation of the CPTs, thereby allowing for practical greedy algorithms that trade off the error and the computational gain derived from the simplification of the network. Tree representations turn out to be particularly suitable in this regard. In particular, we show that decision-tree construction algorithms from the machine learning community can be used to construct an appropriate CPT-tree from a full conditional probability table; pruning algorithms [17] can then be used on this tree, or on one acquired directly from the user, to simplify the CPT-tree in order to allow for faster inference.

Structured representation of CPTs have also proven beneficial in learning Bayesian networks from data [5]. Due to the compactness of the representation, learning procedures are capable of inducing networks that better emulate the true complexity of the interactions present in the data.

This paper represents a starting point for a rigorous extension of Bayesian network representations to incorporate context-specific independence. As we have seen, CSI has a deep and far-ranging impact on the theory and practice of many aspects of probabilistic inference, including representation, inference algorithms, approximation and learning. We consider the exploration and development of these ideas to be a promising avenue for future research.

**Acknowledgements:** We would like to thank Dan Geiger, Adam Grove, Daishi Harada, and Zohar Yakhini for useful discussions. Some of this work was performed while Nir Friedman and Moises Goldszmidt were at Rockwell Palo Alto Science Center, and Daphne Koller was at U.C. Berkeley. This work was supported by a University of California President's Postdoctoral Fellowship (Koller), ARPA contract F30602-95-C-0251 (Goldszmidt), an IBM Graduate fellowship and NSF Grant IRI-95-03109 (Friedman), and NSERC Research Grant OGP0121843 (Boutilier).